**Economic Anthropology in the Era of Generative Artificial Intelligence**

Zachary Sheldon, Peeyush Kumar

2024

**Why Economic Anthropology and GenAI?**

How does a machine see the data of economics?[1] Large Language Models [LLMs] use vast data sets to make predictions about human communication. And economic development agencies use specially trained LLMs to simulate human decision making. Our intervention attempts to address inductive biases in AI research by going beyond the criticism that these models are "only as good as their data." Both human and inhuman intelligences require categories of understanding that weigh attentions towards data *that count as* examples (or "tokens") of a particular type.[2] So, before LLMs can take up more diverse data, they need an ontological update on what types of activity count as "economic." Economic anthropologists have documented the linguistic categories and symbolic practices whereby human beings realize alternative ontologies of value as diverse forms of life.[3] They have also shown how the capitalist market subsumes plural ontologies under the laws of capitalist commodity exchange.[4] <u>How can LLMs help make these non-market economic systems visible and sustainable?</u>

**<u>Methods: How are GenAI models "enculturated"?</u>**

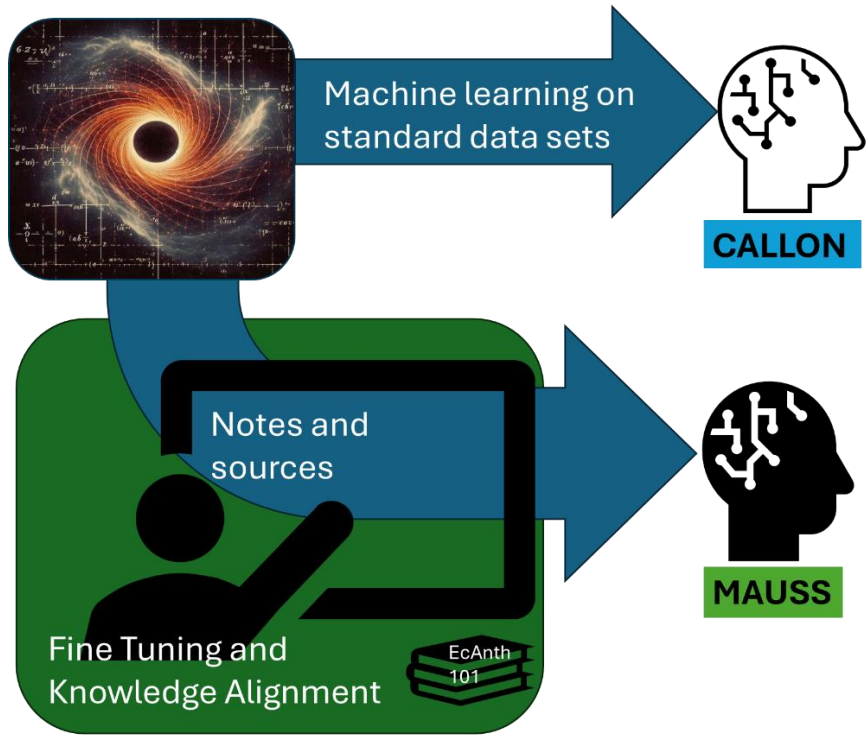

*Figure 1 Two Models for Comparative Analysis*

To explore how research on non-market economies can help LLMs recognize alternative economies, we created two AI "siblings", a control model named **C.A.L.L.O.N.**, ("**C**onventionally **A**verage **L**ate **L**iberal **ON**tology") and a modified twin called **M.A.U.S.S.** ("**M**ore **A**ccurate **U**nderstanding of **S**ociety and its **S**ymbols.")[5] CALLON received only the standard training data. MAUSS underwent fine tuning and knowledge alignment using a "syllabus" that drew from structuralist, Marxist, and feminist works, all of which conceptualize "the economy" as a symbolically-mediated system of activities, values, and agents functionally integrated for the reproduction of collective life.[6] Special weight was given to studies of Trobriand Kula exchange and Balinese Water Temples, which we felt exemplified advances and debates around making non-market economies visible.

**Results: What Does Anthropological Intelligence Add to GenAI?**

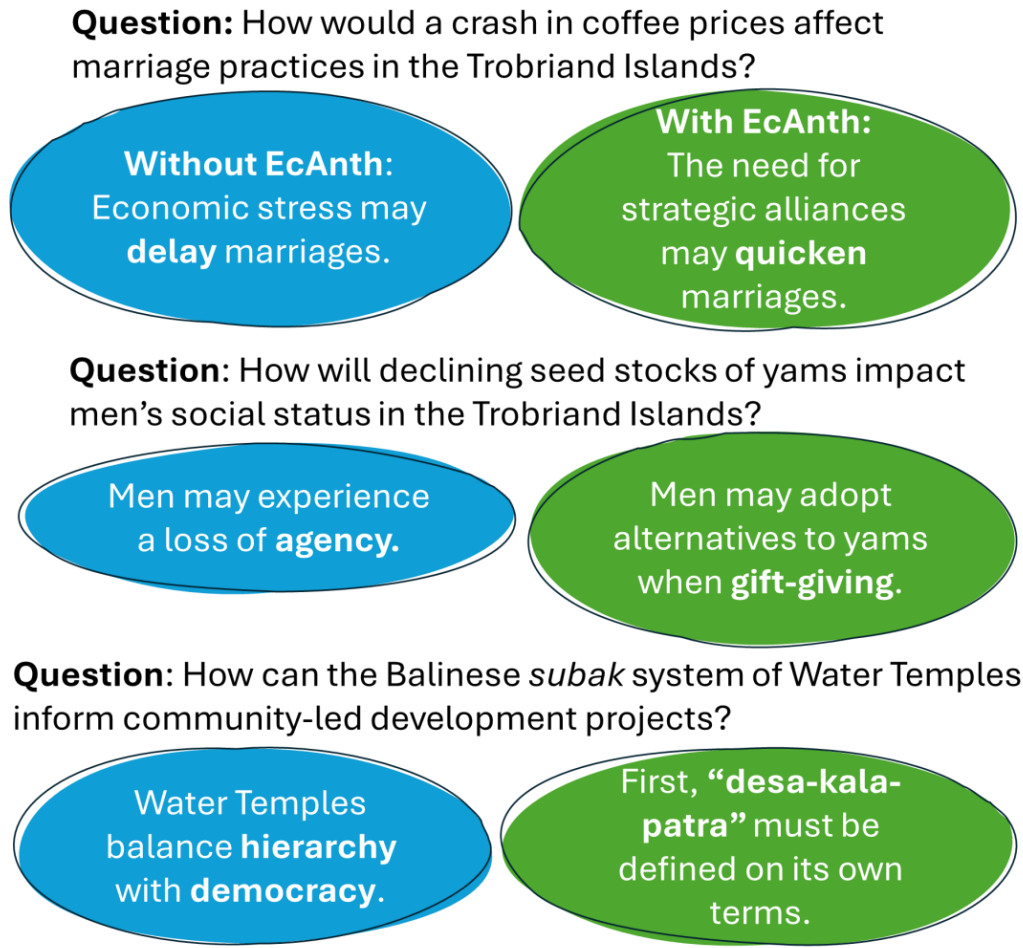

Figure 2 Comparative Analysis: Results

We prompted the two models with questions about climate change and sustainable development. The models <u>do not</u> have any grounded sense of context. And their "hallucinated" responses <u>should not</u> be taken as epistemic claims about real people. But, as "toy models"[7], the pair do show how Economic Anthropology training enhanced an LLM's ability to recognize more plural concepts of "economics."

**Analysis:** The responses highlight three main points of contrast.

- CALLON stressed the fragility of non-market systems, judged acts of social creativity to be a net liability, and invoked a liberal political vocabulary of personal agency.
- Conversely, MAUSS centered the adaptability of non-market economies, recognized the value of socially reproductive activities
- MAUSS was more ready to think with emic concepts of relational autonomy.

**Notes**

1. The following sources were used to produce this collage of economic diversity, read from the top row, left to right.:

   a. https://www.investopedia.com/ask/answers/09/traders-floor-exchange.asp

   b. https://library.panos.co.uk/stock-photo-yam-harvest-festival-women-stacking-yams-in-village-compound-kiriwina-panos-image00015064.html

   c. https://www.britannica.com/money/supply-and-demand

   d. https://notes-culture.blogspot.com/2017/08/levi-strauss-problem-of-avunculate.html

   e. https://www.middleeasteye.net/news/factory-tomorrow-egypt-aims-replace-far-east-cheap-clothing-king

   f. https://en.wikipedia.org/wiki/Florence_Owens_Thompson

   g. https://www.cadtm.org/Anti-Debt-Coalition-Indonesia

   h. https://lenjourneys.com/subak-water-system-bali/

   i. http://www.aboutsanteria.com/blogs/when-a-santeroa-dies

2. For all our differences, both human and inhuman "discursive apperceptive intelligences" (Negarestani 2018) encounter reality through the mediation of conceptual categories that enable us to parse and organize the data of experience. For this reason, although the content of our project was focused on "economics", it has involved a major detour into linguistic anthropology, so that we can more generally understand how the generative power of human language, experience, and social creativity can meaningfully inform the design of Generative Artificial Intelligences. We can start by noting that machine models need a lot more help than human children when discerning useful linguistic-cum-ontological typologies because machine models do not possess the fundaments of continuous bodily being in material spacetime that anchors humans' linguistic self-consciousness in tokens of personal deixis. By this, we mean the foundational insight of human, language-using children that the first-person pronoun of which you speak is not the first-person pronoun of which I speak, and, following from this, that the denotational meaning of the token "I" is only discernable via a self-sensing judgement of the type of speaker uttering the token: "I", meaning me, or "I", meaning you (Benveniste 1966). Lacking the "I=I*" principle of discursive self-consciousness (Negarestani 2018), GENAI's technical process for picking out relevant tokens of experience according to criteria of a certain type can be understood, by analogy, to the eternal "third person" narrator of a modern novel, where the "first person" author is nothing more than an artifact of a discourse that reflects an entirely desubjectified social reality. Novels that lack a centering subject can sometimes nonetheless manage to develop and convey their weighty themes through what we might call, by analogy to GENAI, a feed-forward process of embedding and refining a progressively more convincing portrait of socio-linguistically calibrated discourse. In both cases, "[Type/token interdiscursivity] in effect draws the two or more discursive occasions together within the same chronotropic frame, across which discourse seems to 'move' from originary to secondary occasion, no matter whether 'backwards' or 'forward' in orientation within the frame." (Silverstein 2006, 6) Let us keep this image of extreme, but bridgeable, opposites in mind as we begin to search for a point of convergence between, on the one side, the pure self-consciousness of a just-barely-discursive human being, and, on the other, the absolutely inhuman artefact of a fully elaborated universe of discourse. What unity might exist between us and them? Whether discursive apperception is achieved through embodied self-consciousness or not, it appears that any two points along the human/inhuman spectrum require the cognitive capacity to intuit invariant, typifying structures as they are immanently realized in temporally unfolding token-level events. This talent for picking invariant types out of an incoming stream of tokens enables us, and by us we mean all discursive intelligences, human and inhuman, to transform the inchoate flux of sensory inputs into the data of linguistic experience. That being said, once we recognize what human and inhuman intelligences share, we also must acknowledge the very different route they take to achieve this feat. In humans, an artificial element of culture structures even the most seemingly natural and primary linguistic experiences of phonemic and tonal apperception (Sapir 1925). Our human method for correctly "tokenizing" units of language of any size, ranging from the syllables of a phonemic system to the verse structure of an epic poem, involves accumulating the unfolding, temporalized patterns of sense data within our own spatiotemporally continuous consciousness of mind-body being (Jakobson 1960, 364). Via this rather mysterious transformer architecture, we are able to draw analogies between the inner sense of time consciousness, as it is being altered by its encounter with the aesthetic phenomenon, and the outer sense of a persistent, intersubjectively sensible universe, and, what's more, we are able to "project" (in Jakobson's sense) this inner, temporalized organization of aesthetic impressions onto the outer structuring field of linguistic, social, and material relations (or visa versa). To help visualize this idea, Roman Jakobson offered the analogy of two knitting needles – syntax and semantics – weaving the cloth of language. And Luisa Muraro has developed Jakobson's textile analogy of the "interdependence of metaphor and metonymy" into a critique of many social and natural scientists' one-sided tendency to privilege the semantic, metaphoric, and definitional over the syntactic, metonymic, and combinatorial (cited in Ferme 2001, 10). But, because GENAI lacks the human ability to recursively embed a plurality of voices within a resonate self, it tends to produce a monotonous, Standard Average register that should be familiar to any school essay grader nowadays. That said, GENAI has made significant advance over older text generation programs insofar as its own circuits are modeled on the all too human principle of "self-attention." Much like humans, GENAI's seem to chew over their thoughts, compare them, let them echo aGenAInst each other in the vast and stochastic realm of a theoretically infinite universe of discourse. And, in a process analogous to that of human beings, GENAI models can now perceive relations of relative proximity within the spacetime organization of aesthetic data, then "transform" those patterned events into a judgement about the atemporal, invariant type-level structures that (by statistical abduction) must have produced the token-level patterns. Like Jakobson's Serbian poets, GENAI's self-attention understands that the proximity between any two phonemes, lexemes, or larger discursive chunks within a bounded frame of space and time is indexically relevant to the generation and implementation of higher-level structural invariants that would enable the recreation of those frames in other, self-authored incidents of discursive interaction. But, unlike humans, with our fabulously powerful (although hallucination prone) mind/body memories, GENAI more often fails at "grounding" these transformation/projection/creation operations according to the sensible qualae and pragmatic orders of indexicality afforded by the material environment, by social structures and hierarchies, and indeed by the physical substrate of verbal and non-verbal communication itself. Not only are these factors always in flux, but it is unfortunately the case that the unspoken in culture, that which cannot be said under the existing order, is often where we have the most to learn about the social biases and hierarchies, histories and presents of violence that states and corporations do very often manage to block from circulation within our online discourse, making this dark matter of the social unconscious utterly invisible to GENAI. For without a grounded being in the world, and without any ability to develop consciousness of materially concrete, socially produced ills such as strain, hunger, injury, or even loneliness, the GPT engine "hallucinates" within its own echo chamber, like a talkative bather in a sensory deprivation tub, trying to fill in answers based on potentiality and imagination as it does not know what is quite right. This is equivalent to an interviewee rambling potential answers when they do not have confidence in their answer. Our own human conversations do not lapse into endless games of word association because we are able to limit the arbitrariness of our sign systems by internalizing indexical relationships through the residual consciousness of sensuous impressions – knowing when to end the game of talk in the same way an improvising duet of musicians seem to know when it's time to pick up another tune. And this rhythmic interplay of shared memory is what enables us to alternatively compress, re-access, and edit spacetime-dependent orders of intelligibility by artificializing them as aesthetic/semiotic forms (myth, ritual, games, poetry, music, voicing, flavor, and, in a word, "culture." ) Hence, Levi-Strauss' claim that mythology and music generate the subjectively hallucinatory but cognitively necessary sensation of timelessness: "Because of the level of internal organization of the musical work, the act of listening to it immobilizes passing time; it catches and enfolds it as one catches and enfolds a cloth flapping in the wind." And because our own transformer architecture is not bound by the entropic constraints of asymmetric time, we can enculturate new human intellects without the energy-intensive process of "generating" intelligence from scratch for each new "model". Instead, most of us humans seem to recognize and explicitly thematize in our own rituals of birthing, growing, mating, and dying that the generative logics of culture and language have been borrowed from the already-given intelligibility of our world, which many of us depict in the form of "metahuman" spirit beings (Sahlins 2022). As metahumans are to humans, so we are to GENAI, providing the freely given (and so far, uncompensated) added value of our Nature as the thinking material of their Culture. Human minds, it turns out, are good for computers to think with. Reversing the technological alienation of our generative intellects begins with the recognition that the living labor of human being is the sole and complete source of GENAI's value, in part because these models learn from our data, but also because the ability to parse and organize that data has been copied off our own socio-semiotic protocols. And this fact makes human trainers essential for helping a plurality of GENAIs produce better representations of value production, defined as socio-material constellations of agency oriented towards the ongoing recreation/preservation of aesthetic/material forms upon which can be grounded the qualisigns of value that a community regards as essential for a meaningful, collectively self-conscious existence (Munn 1992).

3. Graeber 2001.

4. Sahlins 1994.

5. Michel Callon replaced the conceptual question of "What is the economy?" with the empirical question of "Where is the market?", leading him to assert that the social scientific abstraction that we call "the economy" is nothing other than a "format" that was performed into existence by postwar Western economists (1988, 2). To model Callon's position in the form of an LLM, one need only train the model on the already available textual corpus of post-industrial Western capitalism, given that any performed instance of the discourse token "economics" or "the economy" will form the statistically average center of attention for an associative network of other, token-level terms. However, although token-level linguistic performatives of the kind described by Callon have played a historically outsized role in the economies of capitalist states, the power of the performative token does not exhaust the concept of economics as a field of human "social creativity", understood as the linguistically/symbolically mediated, historically/mythologically self-consciousness agency of intelligent beings conceptualizing and transforming their own conditions of existence (Graeber 2012). Marcel Mauss, on the other hand, acknowledged the formal autonomy of generative exchange as an existentially human practice that took up various "forms and reasons" across different cases, opening the possibility for theorizing type-level conceptual distinctions based on their functional parallelism across diverse societies, and, in Mauss's own radical argument, even identifying deficiencies in the dominant form of exchange from the perspective of non-dominant forms. Insofar as reflective attention to ethnographic type-tokens like kula, potlach, or mana enhances human economic anthropologists' capacity to recognize patterns of value-creation and transformation within any new set of ethnographic data, a Maussian methodology can meaningfully inform the mechanics of machine learning and provide a touchstone for the integration of anthropological knowledge with AI research. In a future publication, Sheldon will elaborate on this contrast between the "flat ontology" of Actor Network Theory and the "depth ontology" that continues to be generatively employed by logicians, mathematicians, and computer scientists (as well as mystics, magicians, and illusionists), both ancient and modern.

6. The selection of texts followed (Bear et. al. 2015)'s feminist centering of what we might call generativity as a way of "challeng[ing] discursive representations of 'the economic' as a domain."

7. "Toy models are simplified or compressed models that are capable of accommodating a wide range of theoretical assumptions for the purpose of organizing and constructing overarching narratives (or explicit metatheories) that change the standard and implicit metatheoretical interpretations according to which such theoretical items are generally represented" (Negarestani 2018. 124).